\documentclass[runningheads]{llncs}
\usepackage{graphicx}
\usepackage{amsmath}
\usepackage{amsfonts}
\usepackage{amssymb}
\usepackage{mathtools}
\usepackage{multirow}
\usepackage{units}
\usepackage{array}
\usepackage{diagbox}
\usepackage{rotating}
\usepackage{siunitx}
\usepackage{etoolbox}
\usepackage{booktabs}
\usepackage{slashbox}
\usepackage{hhline}
\usepackage{footmisc}
\usepackage{stfloats}
\usepackage{algorithmic}
\usepackage{caption}

\newcolumntype{H}{>{\setbox0=\hbox\bgroup}c<{\egroup}@{}}

\usepackage[linesnumbered,ruled]{algorithm2e}

\begin{document}
\title{Graph Classification with 2D Convolutional Neural Networks}
%
%
\author{Antoine J.-P. Tixier*\inst{1} \and
Giannis Nikolentzos\inst{1} \and
Polykarpos Meladianos\inst{2} \and
Michalis Vazirgiannis\inst{1,2}}
\authorrunning{A. Tixier et al.}

\institute{\'Ecole Polytechnique, France \\
\email{\{anti5662,nikolentzos,mvazirg\}@lix.polytechnique.fr} \and
Athens University of Economics and Business, Greece \\
\email{pmeladianos@aueb.gr} \\
*corresponding author}
\maketitle              
\begin{abstract}
Graph learning is currently dominated by graph kernels, which, while powerful, suffer some significant limitations. Convolutional Neural Networks (CNNs) offer a very appealing alternative, but processing graphs with CNNs is not trivial. To address this challenge, many sophisticated extensions of CNNs have recently been introduced. In this paper, we reverse the problem: rather than proposing yet another graph CNN model, \textit{we introduce a novel way to represent graphs as multi-channel image-like structures that allows them to be handled by vanilla 2D CNNs}. 
Experiments reveal that our method is more accurate than state-of-the-art graph kernels and graph CNNs on 4 out of 6 real-world datasets (with and without continuous node attributes), and close elsewhere. Our approach is also preferable to graph kernels in terms of time complexity. Code and data are publicly available\footnote{\scriptsize{https://github.com/Tixierae/graph\_2D\_CNN}}.
\end{abstract} 

\section{Introduction}

\noindent \textbf{Graph classification}. 
Graphs, or networks, are rich, flexible, and universal structures that can accurately represent the interaction among the components of many natural and human-made complex systems. A central graph mining task is that of \textit{graph} classification (not to be mistaken with \textit{node} classification). The instances are full graphs and the goal is to predict the category they belong to. The applications of graph classification are numerous and range from determining whether a protein is an enzyme or not in bioinformatics, to categorizing documents in NLP, and social network analysis. Graph classification is the task of interest in this study.

\noindent \textbf{Limitations of graph kernels}.
The state-of-the-art in graph classification has traditionally been dominated by \textit{graph kernels}. Graph kernels compute the similarity between two graphs as the sum of the pairwise similarities between some of their substructures, and then pass the similarity matrix computed on the entire dataset to a kernel-based supervised algorithm such as the Support Vector Machine \cite{cortes1995support} to learn soft classification rules. Graph kernels mainly vary based on the substructures they use, which include random walks \cite{gartner2003graph}, shortest paths \cite{borgwardt2005shortest}, and subgraphs \cite{shervashidze2009efficient}, to cite only a few. While graph kernels have been very successful, they suffer significant limitations:

\textit{L1: High time complexity}. This problem is threefold: first, populating the kernel matrix requires computing the similarity between every two graphs in the training set (say of size $N$), which amounts to $\nicefrac{N(N-1)}{2}$ operations. The cost of training therefore increases much more rapidly than the size of the dataset. Second, computing the similarity between a pair of graphs (i.e., performing a single operation) is itself polynomial in the number of nodes. For instance, the time complexity of the shortest path graph kernel is $\mathcal{O} (|V_{1}|^2|V_{2}|^2)$ for two graphs $(V_{1},V_{2})$, where $|V_{i}|$ is the number of nodes in graph $V_{i}$. Processing large graphs can thus become prohibitive, which is a serious limitation as big networks abound in practice. Finally, finding the support vectors is $\mathcal{O}(N^2)$ when the $C$ parameter of the SVM is small and $\mathcal{O}(N^3)$ when it gets large \cite{bottou2007support}, which can again pose a problem on big datasets.

\textit{L2: Disjoint feature and rule learning}. With graph kernels, the computation of the similarity matrix and the learning of the classification rules are two independent steps. In other words, the features are fixed and not optimized for the task.

\textit{L3: Graph comparison is based on small independent substructures}. As a result, graph kernels focus on local properties of graphs, ignoring their global structure \cite{nikolentzos2017matching}. They also underestimate the similarity between graphs and suffer unnecessarily high complexity (due to the explosion of the feature space), as substructures are considered to be orthogonal dimensions \cite{yanardag2015deep}.

\begin{figure*}[h]
  \centering
    \includegraphics[width=0.95\textwidth]{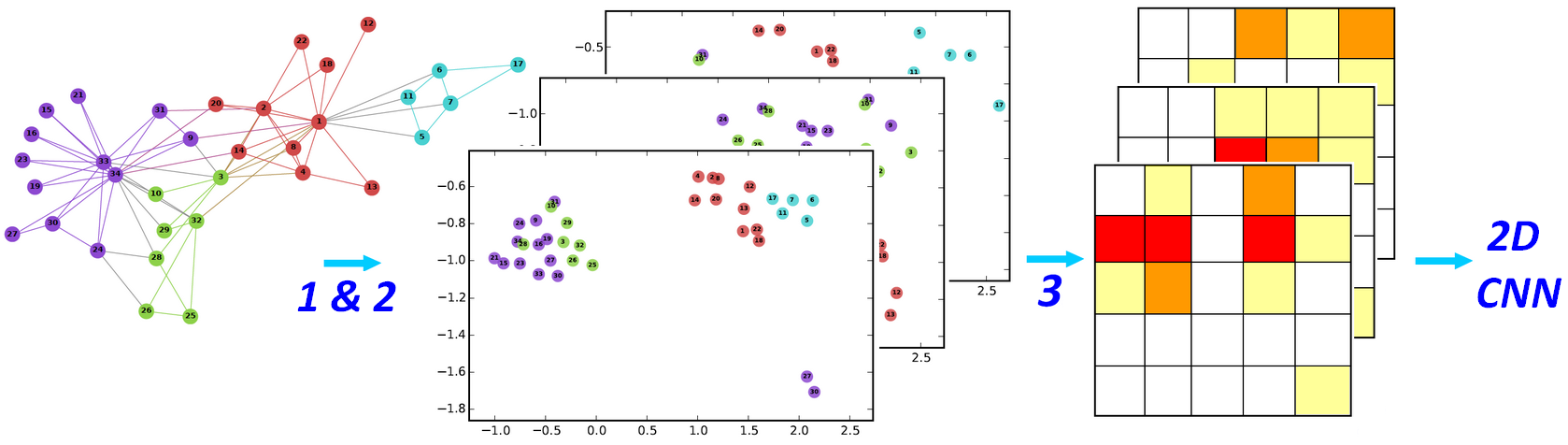}
\captionsetup{justification=justified,size=small}
    \caption{Our 3-step approach represents graphs as ``images'' suitable for vanilla 2D CNNs. Continuous node attribute vectors can be passed as extra channels. Steps 1 \& 2: graph node embeddings and compression with PCA. Step 3: computation and stacking of the 2D histograms.}
\label{fig:overview}
\end{figure*}

\vspace{-1cm}

\section{Proposed method}
\noindent \textbf{Overview}.
We propose a simple approach to turn a graph into a multi-channel image-like structure suitable to be processed by a traditional 2D CNN. The process (summarized in Fig. \ref{fig:overview}) can be broken down into 3 steps:
(1) graph node embedding,
(2) embedding space compression,
(3) repeated extraction of 2D slices from the compressed space and computation of a 2D histogram for each slice.

The ``image'' representation of the graph is finally given by the stack of its 2D histograms (each histogram making for a channel). Note that the dimensionality of the final representation of a graph does not depend on its number of nodes or edges. Big and small graphs are represented by images of the same size.

Our method addresses the limitations of graph kernels in the following ways: 

\textit{L1}. By converting all graphs in a given dataset to representations of the same dimensionality, and by using a classical 2D CNN architecture for processing those graph representations, our method offers \textit{constant time} complexity at the instance level, and \textit{linear time} complexity at the dataset level. Moreover, state-of-the-art node embeddings can be obtained for a given graph in \textit{linear time} w.r.t. the number of nodes in the graph, for instance with \texttt{node2vec} \cite{grover2016node2vec}.

\textit{L2}. Thanks to the 2D CNN classifier, features are learned directly from the raw data during training such that classification accuracy is maximized.

\textit{L3}. Our approach capitalizes on state-of-the-art graph node embedding techniques that capture both local and global properties of graphs. In addition, we remove the need for handcrafted features.

\noindent \textbf{How to represent graphs as structures that verify the spatial dependence property?} Convolutional Neural Networks (CNNs) are feedforward neural networks specifically designed to work on regular grids \cite{lecun1998gradient}. A regular grid is the $d$-dimensional Euclidean space discretized by parallelotopes (rectangles for $d=2$, cuboids for $d=3$, etc.). 
Regular grids satisfy the \textit{spatial dependence}\footnote{\scriptsize{the concept of spatial dependence is well summarized by:
``everything is related to everything else, but near things are more related than distant things'' \cite{tobler1970computer}. For instance in images, close pixels are more related than distant pixels.}} property, which is the fundamental premise on which local receptive fields and hierarchical composition of features in CNNs hold.

\vspace{-0.2cm}

\begin{figure*}[h]
  \centering
    \includegraphics[width=0.95\linewidth]{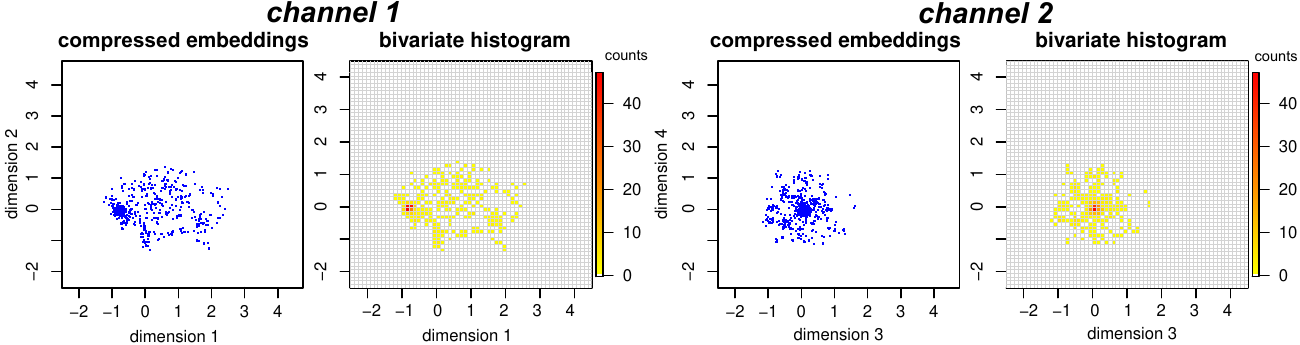}
\captionsetup{justification=justified,singlelinecheck=false, size=small}
    \caption{Node embeddings and image representation of graph ID \#10001 (577 nodes, 1320 edges) from the REDDIT-12K dataset.}
    \label{fig:hist_scatter}
\end{figure*}

\vspace{-0.4cm}

Traditionally, a graph $G(V,E)$ is encoded as its adjacency matrix $A$ or Laplacian matrix $L$. $A$ is a square matrix of dimensionality $|V| \times |V|$, symmetric in the case of undirected graphs, whose $(i,j)^{th}$ entry $A_{i,j}$ is equal to the weight of the edge $e_{i,j}$ between nodes $v_{i}$ and $v_{j}$, if such an edge exists, or to 0 otherwise. On the other hand, the Laplacian matrix $L$ is equal to $D-A$, where $D$ is the diagonal degree matrix. One could initially consider passing one of those structures as input to a 2D CNN. However, unlike in images, where close pixels are more strongly correlated than distant pixels, adjacency and Laplacian matrices are not associated with spatial dimensions and the notion of Euclidean distance, and thus do not satisfy the spatial dependence property. As will be detailed next, we capitalize on \textit{graph node embeddings} to address this issue.

\underline{Step 1: Graph node embeddings}. There is local correlation in the node embedding space. In that space, the Euclidean distance between two points is meaningful: it is inversely proportional to the similarity of the two nodes they represent. For instance, two neighboring points in the embedding space might be associated with two nodes playing the same structural role (e.g., of flow control), belonging to the same community, or sharing some other common property.

\underline{Step 2: Alignment and compression with PCA}. As state-of-the-art node embedding techniques (such as \texttt{node2vec}) are neural, they are stochastic. Dimensions are recycled from run to run, which means that a given dimension will not be associated with the same latent concepts across graphs, or across several runs on the same graph. Therefore, to ensure that the embeddings of all the graphs in the collection are comparable, we apply PCA and retain the first $d \ll D$ principal components (where $D$ is the dimensionality of the original node embedding space). PCA also serves an information maximization (compression) purpose. Compression is desirable in terms of complexity, as it greatly reduces the shape of the tensors fed to the CNN (and thus the number of channels, for reasons that will become clear in what follows), at the expense of a negligible information loss.

\underline{Step 3: Computing and stacking 2D histograms}. We finally repeatedly extract 2D slices from the $d$-dimensional PCA node embedding space, and turn those planes into regular grids by discretizing them into a finite, fixed number of equally-sized bins, where the value associated with each bin is the count of the number of nodes falling into that bin. In other words, we represent a graph as a stack of $d/2$ 2D histograms of its (compressed) node embeddings\footnote{\scriptsize{our representation is unrelated to the widespread \textit{color histogram} encoding of images.}}. As illustrated in Fig. \ref{fig:hist_scatter}, the first histogram is computed from the coordinates of the nodes in the plane made of the first two principal directions, the second histogram from directions 3 and 4, and so forth. Note that using adjacent and following PCA dimensions is an arbitrary choice. It ensures at least that channels are sorted by decreasing order of informativeness.

Using computer vision vocabulary, bins can be viewed as \textit{pixels}, and the 2D slices of the embedding space as \textit{channels}. However, in our case, instead of having 3 channels (R,G,B) like with color images,  we have $d/2$ of them. That is, each pixel (each bin) is associated with a vector of size $d/2$, whose entries are the counts of the nodes falling into that bin in the corresponding 2D slice of the embedding space. Finally, the \textit{resolution} of the image is determined by the number of bins of the histograms, which is constant for a given dataset across all channels.

\section{Experiments}

\begin{figure}[h]
  \centering
    \includegraphics[width=0.9\linewidth]{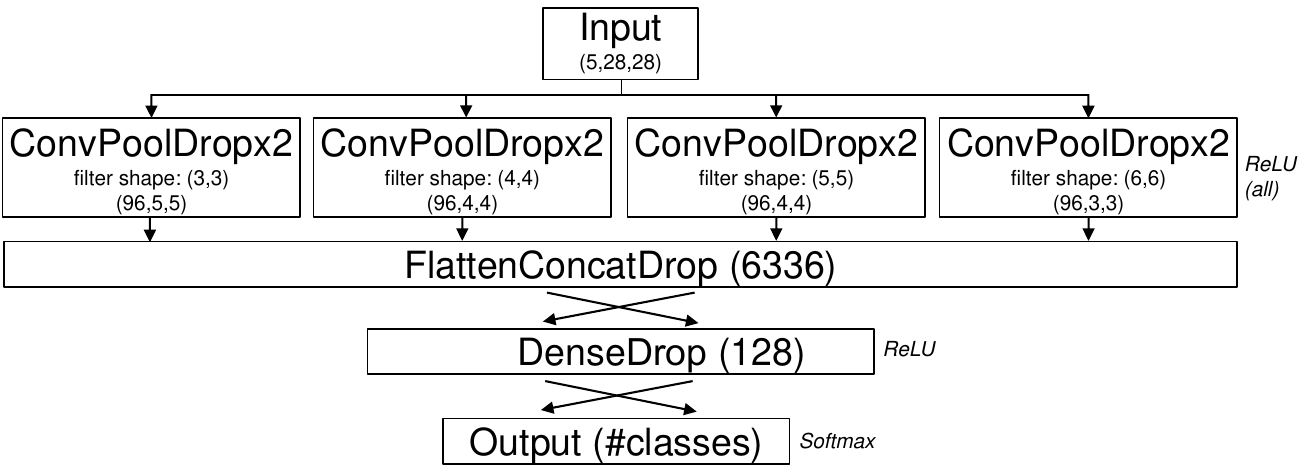}
\captionsetup{justification=justified,singlelinecheck=false, size=small}
    \caption{2D CNN architecture used in our experiments. The number within parentheses refer to the \textit{output} dimensions of the tensors. }
\label{fig:arch}
\end{figure}

\noindent \textbf{2D CNN Architecture}.
We implemented a variant of LeNet-5 \cite{lecun1998gradient} with which we reached 99.45\% accuracy on the MNIST handwritten digit classification dataset. As illustrated in Fig. \ref{fig:arch} for an input of shape (5,28,28), this simple architecture deploys four convolutional-pooling layers (each repeated twice) in parallel, with respective region sizes of 3, 4, 5 and 6, followed by two fully-connected layers. Dropout \cite{srivastava2014dropout} is employed for regularization at every hidden layer. The activations are \texttt{ReLU} functions (in that, our model differs from LeNet-5), except for the ultimate layer, which uses a \texttt{softmax} to output a probability distribution over classes. For the convolution-pooling block, we employ 64 filters at the first level, and as the signal is halved through the (2,2) max pooling layer, the number of filters in the subsequent convolutional layer is increased to 96 to compensate for the loss in resolution.

\vspace{-0.4cm}

\vspace{-0.6cm}

\noindent \textbf{Random graphs}.
To quickly test the viability of our pipeline, we created a synthetic 5-class dataset containing 2000 undirected and unweighted networks in each category. For the first and second classes, we generated graphs with the Stochastic Block Model, featuring respectively 2 and 3 communities of equal sizes. The in-block and cross-block probabilities were respectively set to 0.1 and 0.7, and the size $|V|$ of each graph was randomly sampled from the Normal distribution with mean 150 and standard deviation 30, i.e., $\mathcal{N}(150,30)$. The third category was populated with scale-free networks, that is, graphs whose degree distributions follow a power law, using the Barab{\'a}si-Albert model. The number of incident edges per node was sampled for each graph from $\mathcal{N}(5,2)$, and the size of each graph was drawn from $\mathcal{N}(150,30)$ like for the first two classes. Finally, the fourth and fifth classes were filled with Erd{\"o}s-R{\'e}nyi graphs whose sizes were respectively sampled from $\mathcal{N}(300,30)$ and $\mathcal{N}(150,30)$, and whose edge probabilities were drawn from $\mathcal{N}(0.3,0.15)$. This overall, gave us a large variety of graphs.

\textit{Spectral embeddings}. We started with the most naive way to embed the nodes of a graph, that is, using the eigenvectors of its adjacency matrix (we also experimented with the Laplacian but observed no difference). Here, no PCA-based compression was necessary. We retained the eigenvectors associated with the 10 largest eigenvalues in magnitude, thus embedding the nodes into a $10$-dimensional space.

\textit{Image resolution and channels}. As we computed the unit-normed eigenvectors, the coordinates of any node in any dimension belonged to the $[-1,1]$ range. Furthermore, inspired by the MNIST images which are 28 $\times$ 28 in size, and on which we initially tested our 2D CNN architecture, we decided to learn 2D histograms featuring 28 bins in each direction. This gave us a resolution of $\nicefrac{28}{(1-(-1))}$, that is, 14 pixels per unit, which we write 14:1 for brevity in the remainder of this paper. Finally, we decided to make use of the information carried out by all 10 eigenvectors, and thus kept 5 channels. Any graph in the dataset was thus represented as a tensor of shape (5,28,28).

\begin{figure*}[h]
  \centering
    \includegraphics[width=0.85\textwidth]{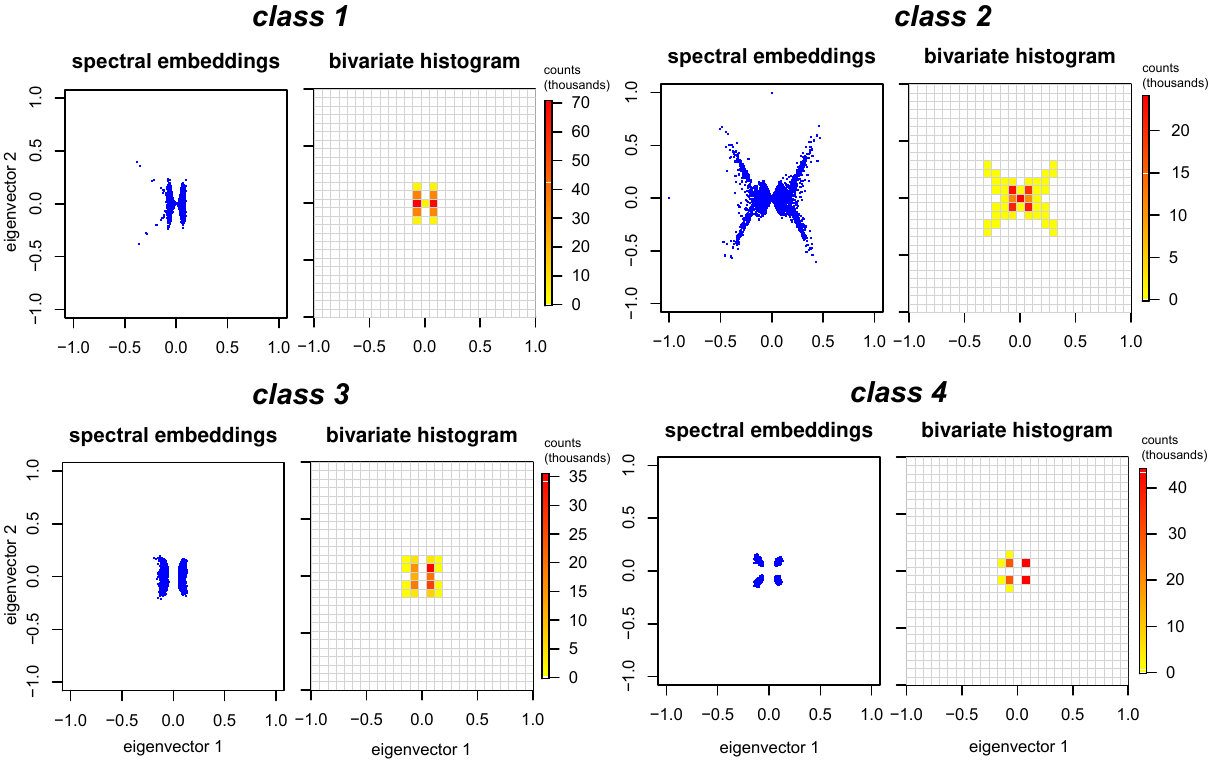}
\captionsetup{justification=justified,singlelinecheck=false, size=small}
    \caption{Overlayed node embeddings in the space made of the first two eigenvectors of the adjacency matrices (first channel) and associated overlayed bivariate histograms for all graphs in the first four classes of the \textit{random} dataset.}
    \label{fig:mot_ex}
\end{figure*}

\textit{Results}. In a 10-fold cross validation setting where each fold was repeated 3 times, we reached a mean classification accuracy of $99.08\% (\pm 3.21)$, usually within 3 epochs, which is a very good performance. Even though we injected some variance, categories are associated with specific patterns (see Fig. \ref{fig:mot_ex}) which are easily captured by the CNN.

\vspace{-0.4cm}

\vspace{-0.65cm}

\noindent \textbf{Real-world graphs}.
We conducted experiments on 6 real-world datasets whose statistics are summarized in Table \ref{tab:stats}. In all datasets, graphs are unweighted, undirected, with unlabeled nodes, and the task is to predict the class they belong to. Classes are mutually exclusive. The first five datasets, REDDIT-B, REDDIT-5K, REDDIT-12K, COLLAB, and IMDB-B, come from \cite{yanardag2015deep} and contain social networks. The sixth dataset, PROTEINS\_full, is a bioinformatics one and comes from \cite{KKMMN2016,borgwardt2005protein}. In this dataset, each node is associated with a 29-dimensional continuous vector. To incorporate this extra information into our images, we compressed the attribute vectors with PCA, retaining the same number of dimensions as for the node embeddings, and normalized them to have same range as the node embeddings. Finally, each node was represented by a single vector made of its compressed node embedding concatenated with its compressed and normalized continuous attribute vector. That is, we had $d/2$ channels for node embeddings and $d/2$ channels for node attributes, where $d$ is the number of principal components retained in each case.

\vspace{-0.5cm}

\begin{table*}[h]
\begin{center}
\scalebox{0.75}{
\begin{tabular}{|lccccc|c|} \hline
& IMDB-B & COLLAB & REDDIT-B & REDDIT-5K & REDDIT-12K & PROTEINS\_full \\ \hline\hline
Max \# vertices & 136 & 492 & 3782 & 3648 & 3782 & 620\\
Min \# vertices & 12 & 32 & 6 & 22 & 2 & 4\\
Average \# vertices & 19.77 & 74.49 & 429.61 & 508.50 & 391.40 & 39.05 \\ \hline
Max \# edges & 1249 & 40120 & 4071 & 4783 & 5171 & 1049\\
Min \# edges & 26 & 60 & 4 & 21 & 1 & 5\\
Average \# edges & 96.53 & 2457.78 & 497.75 & 594.87 & 456.89 & 72.82\\ \hline
\# graphs & 1000 & 5000 & 2000 & 4999 & 11929 & 1113\\ \hline
Average diameter & 1.861 & 1.864 & 9.72 & 11.96 & 10.91 & 11.57 \\ \hline
Average density (\%) & 52.06 & 50.92 & 2.18 & 0.90 & 1.79 & 21.21\\ \hline
\# classes & 2 & 3 & 2 & 5 & 11 & 2\\ \hline
Max class imbalance & 1:1 &  1:3.4 & 1:1 &  1:1 &  1:5 & 1:1.5\\ \hline
\end{tabular}
}
\captionsetup{justification=centering, size=small}
\caption{\label{tab:stats} Statistics of the social network datasets (first 5 columns) and the bioinformatics dataset used in our experiments.}
\end{center}
\end{table*}

\vspace{-1cm}

\textit{Neural embeddings}.
We used the \texttt{node2vec} algorithm \cite{grover2016node2vec} as our node embedding method for the real-world networks, as it was giving much better results than the basic spectral method. \texttt{Node2vec} applies the very fast Skip-Gram language model \cite{mikolov2013efficient} to truncated biased random walks performed on the graph. The algorithm scales linearly with the number of nodes in the network. We used the high performance, publicly available C++ implementation of the authors\footnote{\scriptsize{\url{https://github.com/snap-stanford/snap/tree/master/examples/node2vec}}}.

\textit{Image resolution}. We stuck to similar values as in our initial experiments involving spectral embeddings, i.e., 14 pixels per unit (14:1). We also tried 9 pixels per unit (9:1).

\textit{Image size}. On a given dataset, image size is calculated as the range \newline $|max($coordinates$) - min($coordinates$)| \times  \mathrm{resolution}$, where \texttt{coordinates} are the flattened node embeddings. For instance, on COLLAB with a resolution of 9:1, image size is equal to $37 \times 37$, since $|2.78-(-1.33)| \times 9 \approx 37$.

\textit{Number of channels}. With the $p$ and $q$ parameters of \texttt{node2vec} held constant and equal to 1, we conducted a search on the coarse grid $\big\{$(14,9);(2,5)$\big\}$ to get more insights about the impact of resolution and number of channels (respectively). When using 5 channels, the graphs with less than 10 nodes were removed, because for these graphs, we cannot get a $10$-dimensional node embedding space (there cannot be more dimensions than data points). This represented only a few graphs overall. \\ 
On the PROTEINS\_full dataset, we only experimented with 2 node embeddings channels and 2 node attribute channels.

\textit{p,q,c, and $d_{n2v}$ \texttt{node2vec} parameters}. With the best resolution and number of channels, we then tuned the return and in-out parameters $p$ and $q$ of \texttt{node2vec}. Those parameters respectively bias the random walks towards exploring larger areas of the graph or staying in local neighborhoods, allowing the embeddings to encode a similarity that interpolates between structural equivalence (two nodes acting as, e.g., flow controllers, are close to each other) and homophily (two nodes belonging to the same community are close to each other). Following the \texttt{node2vec} paper \cite{grover2016node2vec}, we tried 5 combinations of values for $(p,q)$: $\big\{(1,1);(0.25,4);(4,0.25);(0.5,2);(2,0.5)\big\}$. Note that $p=q=1$ is equivalent to \texttt{DeepWalk} \cite{perozzi2014deepwalk}.\\
Since the graphs in the COLLAB and the IMDB-B datasets are very dense ($>50\%$, average diameter of 1.86), we also tried to set the context size $c$ to smaller values of 1 and 2 (the default value is $c=10$). The context size is used when generating the training examples (\texttt{context},\texttt{target}) to pass to the skip-gram model: in a given random walk, it determines how many nodes \textit{before} and \textit{after} the target node should be considered part of the context.\\
Finally, $d_{n2v}$ is the dimension of the node embeddings learned by \texttt{node2vec}. The default value is 128. Since the graphs in COLLAB, IMDB-B, and PROTEINS\_full are small (20, 74, and 39 nodes on average), we experimented with lower values than the default one on this dataset, namely 12 and 4. Note that with $d_{n2v}=4$, we can get at most 2 channels. The final values of $p,q,c$, and $d_{n2v}$ for each dataset are summarized in Table \ref{table:best}. 

\vspace{-0.25cm}

\begin{table}[h]
\begin{center}
\scalebox{0.7}{
\begin{tabular}{ |r|ccccc|c| } 
 \hline
 & REDDIT-B & REDDIT-5K & REDDIT-12K & COLLAB & IMDB-B & PROTEINS\_full\\
 \hline
 Res. & 9:1 & 9:1 & 9:1 & 9:1 & 14:1 & 9:1  \\ 
 \#Chann. & 5 & 2 & 5 & 5 & 5 & 2/2$^\star$\\
 p,q & 2,0.5 & 4,0.25 & 1,1 & 0.5,2 & 1,1 & 0.5,2\\
  c,$d_{n2v}$ & - & - & - & 2,12 & - & -,4\\
 \hline
\end{tabular}
}
\end{center}
\captionsetup{justification=centering, size=small}
\caption{Final resolution, number of channels, and $p,q,c$, and $d_{n2v}$ \texttt{node2vec} parameters for each dataset. $^\star$number of channels for node embeddings and continuous node attributes. - means default value(s).}
\label{table:best}
\end{table}

\vspace{-0.8cm}

\textit{Baselines}.
On the social network datasets (on which there are no continuous node attributes), we re-implemented two state-of-the-art graph kernels, the graphlet kernel \cite{shervashidze2009efficient} (we sampled 2000 graphlets of size up to 6 from each graph) and the Weisfeiler-Lehman (WL) kernel framework \cite{shervashidze2011weisfeiler} with the subtree graph kernel \cite{gartner2003graph} (we used node degrees as labels).

We also report for comparison purposes the performance of multiple state-of-the-art baselines that all share with us the same experimental setting: Deep Graph Kernels (DGK) \cite{yanardag2015deep}, the best graph CNN from \cite{niepert2016learning}, namely PSCN $k=10$, and Deep Graph CNN (DGCNN) \cite{zhang2018end}. Note that to be fair, we excluded DGK from the comparison on the bioinformatics dataset since it doesn't make use of node attributes.

On the PROTEINS\_full dataset, we compared with the following baselines, which all take into account node attribute vectors:
Hash Graph Kernel (HGK-SP) with shortest-path base kernel \cite{morris2016faster},
Hash Graph Kernel with Weisfeiler-Lehman subtree base kernel (HGK-WL) \cite{morris2016faster},
the best performing variant of Graph invariant kernels (GIK) \cite{orsini2015graph},
GraphHopper \cite{feragen2013scalable} and baselines within,
propagation kernel with diffusion scheme (PROP-diff) \cite{neumann2012efficient}, and 
propagation kernel with hashing-based label discretization and WL kernel update (PROP-WL) \cite{neumann2012efficient}.

\textit{Configuration}.
Following best practice, we used 10-fold cross validation and repeated each fold 3 times in all our experiments. For the graphlet and WL kernels, we used a C-SVM classifier\footnote{\scriptsize{\url{http://scikit-learn.org/stable/modules/generated/sklearn.svm.SVC.html}}} \cite{scikitlearn}. The C parameter of the SVM and the number of iterations in WL were jointly optimized on a 90-10 \% partition of the training set of each fold by searching the grid $\big\{(10^{-4},10^{4},\mathrm{len}=10);(2, 7, \mathrm{step}=1)\big\}$.

For our 2D CNN, we used Xavier initialization \cite{glorot2010understanding}, a batch size of 32, and for regularization, a dropout rate of 0.3 and early stopping with a patience of 5 epochs (null delta). The categorical cross-entropy loss was optimized with Adam \cite{kingma2014adam} (default settings). We implemented our model in \texttt{Keras} \cite{chollet2015keras} version 1.2.2\footnote{\scriptsize{\url{https://faroit.github.io/keras-docs/1.2.2/}}} with \texttt{tensorflow} \cite{abadi2016tensorflow} backend. The hardware used consisted in an NVidia Titan X Pascal GPU with an 8-thread Intel Xeon 2.40 GHz CPU and 16 GB of RAM. The graph kernel baselines were run on an 8-thread Intel i7 3.4 GHz CPU, with 16 GB of RAM.

\section{Results}
Results are reported in Table \ref{tab:results} for the social network datasets and Table \ref{tab:results_bio} for the bioinformatics dataset.

\vspace{-0.45cm}

\begin{table*}[!h]
\begin{center}
\scalebox{0.78}{
{\small
\begin{tabular}{|l|ccccc|} \hline
\multirow{2}{*}{\backslashbox{Method}{Dataset}} & \multirow{2}{*}{\shortstack{REDDIT-B \\\scriptsize{(size=2,000;nclasses=2)}}} & \multirow{2}{*}{\shortstack{REDDIT-5K \\\scriptsize{(4,999;5)}}} & \multirow{2}{*}{\shortstack{REDDIT-12K \\\scriptsize{(11,929;11)}}} & \multirow{2}{*}{\shortstack{COLLAB \\\scriptsize{(5,000;3)}}} & \multirow{2}{*}{\shortstack{IMDB-B \\\scriptsize{(1,000;2)}}} \\ 
& & & & & \\ \hline \hline
Graphlet \scriptsize{Shervashidze2009} &  77.26 ($\pm$ 2.34) &  39.75 ($\pm$ 1.36) &  25.98 ($\pm$ 1.29) & 73.42 ($\pm$ 2.43) &  65.40 ($\pm$ 5.95) \\
WL \scriptsize{Shervashidze2011} &  78.52 ($\pm$ 2.01) & 50.77 ($\pm$ 2.02) & 34.57 ($\pm$ 1.32) & \textbf{77.82}$^{\star}$ ($\pm$ 1.45) &  \textbf{71.60} ($\pm$ 5.16) \\
Deep GK \scriptsize{Yanardag2015} & 78.04 ($\pm$ 0.39) & 41.27 ($\pm$ 0.18) & 32.22 ($\pm$  0.10) & 73.09 ($\pm$ 0.25) &  66.96 ($\pm$ 0.56 ) \\
PSCN $k=10$ \scriptsize{Niepert2016} & 86.30 ($\pm$ 1.58) &  49.10 ($\pm$ 0.70) &  41.32 ($\pm$ 0.42) & 72.60 ($\pm$ 2.15) &  71.00 ($\pm$ 2.29) \\
DGCNN \scriptsize{Zhang2018} & - &  - &  - & 73.76 ($\pm$ 0.49) & 70.03 ($\pm$ 0.86) \\
2D CNN (our method) & \textbf{89.12}$^{\star}$ ($\pm$ 1.70) & \textbf{52.11} ($\pm$ 2.24) & \textbf{48.13}$^{\star}$ ($\pm$ 1.47) & 71.33 ($\pm$ 1.96) & 70.40 ($\pm$ 3.85) \\
\hline
\end{tabular}
}
}
\captionsetup{justification=justified,singlelinecheck=false, size=small}
\caption{10-fold CV average test set classification accuracy of our proposed method compared to state-of-the-art graph kernels and graph CNNs, on the social network datasets. $\pm$ is standard deviation. Best performance per column in \textbf{bold}. $^{\star}$indicates stat. sign. at the $p<0.05$ level (our 2D CNN vs. WL) using the Mann-Whitney U test.}
\label{tab:results}
\end{center}
\end{table*}

\vspace{-2cm}

\begin{table*}[!h]
\begin{center}
\scalebox{0.75}{
{\small
\begin{tabular}{ |r|ccccccccc| } 
 \hline
 & \textbf{2D CNN} & DGCNN & PSCN $k=10$ & HGK-SP & HGK-WL & GIK & GraphHopper & PROP-diff & PROP-WL \\
 & \textbf{our method} & Zhang18 & Niepert16 & Morris16 & Morris16  & Orsini15 & Feragen13 & Neumann12 & Neumann12 \\
 \hline
 Acc. & \textbf{77.12} & 75.54 & 75.00 & 75.14 & 74.88 & 76.6 & 74.1 & 73.3 & 73.1  \\ 
Std. dev. & 2.79 & 0.94 & 2.51 & 0.47 &  0.64 & 0.6 & 0.5 & 0.4 & 0.8 \\ 
 \hline
\end{tabular}
}
}
\end{center}
\captionsetup{justification=centering, size=small}
\caption{10-fold CV average test set classification accuracy of our proposed method compared to state-of-the-art graph kernels and graph CNNs on the bioinformatics dataset (PROTEINS\_full).}
\label{tab:results_bio}
\end{table*}

\vspace{-1cm}

Our approach shows statistically significantly better than all baselines on the REDDIT-12K and REDDIT-B datasets, with large absolute improvements of \textbf{6.81} and \textbf{2.82} in accuracy over the best performing competitor, respectively.
We also reach best performance on the REDDIT-5K and PROTEINS\_full datasets, with respective improvements in accuracy of \textbf{1.34} and \textbf{0.52} over the best performing baselines. In particular, the fact that we reach best performance on PROTEINS\_full shows that our approach is flexible enough to leverage not only the topology of the network, but also continuous node attributes, in a very simple and unified way (we simply concatenate node embeddings with node attribute vectors). Note that when not using node attributes (only the first 2 node embeddings channels), the performance of our model decreases from 77.12 to 73.43, showing that our approach is capable of using both topological and attribute information.

Finally, on the IMDB-B dataset, we get third place, very close ($\leq$ 1.2) to the top performers (no statistically significant difference).
The only dataset on which a baseline proved significantly better than our approach is actually COLLAB (WL graph kernel). On this dataset though, the WL kernel beats \textit{all} models by a wide margin, and we are relatively close to the other Deep Learning approaches ($\leq$ 2.43 difference).

\noindent \textbf{Runtimes}.
Even if not directly comparable, we report in Table \ref{table:time} kernel matrix computation time for the two graph kernel baselines, along with the time per epoch of our 2D CNN model.

\vspace{-0.45cm}

\begin{table*}[!h]
\begin{center}
\scalebox{0.74}{
{\small
\begin{tabular}{ |r|ccccc|c| } 
 \hline
 & REDDIT-B & REDDIT-5K & REDDIT-12K & COLLAB & IMDB-B & PROTEINS\_full \\
 \hline
 \scriptsize{Size, average (\# nodes, \# edges)} & \scriptsize{2000, (430,498)} & \scriptsize{4999, (509,595)} & \scriptsize{11929, (391,457)} & \scriptsize{5000, (74,2458)} & \scriptsize{1000, (20,97)} & 
 \scriptsize{1113, (39,73)} \\
 \scriptsize{Input shapes (for our approach)} & \scriptsize{(5,62,62)} & \scriptsize{(2,65,65)} & \scriptsize{(5,73,73)} & \scriptsize{(5,36,36)} & \scriptsize{(5,37,37)} & \scriptsize{(4,70,70)} \scriptsize{(2,70,70)}$^\star$ \\
 \hline  \hline
 Graphlet \scriptsize{Shervashidze2009} & 551 & 5046 & 12208 & 3238 & 275 & -  \\
 WL \scriptsize{Shervashidze2011} & 645 & 5087 & 20392 & 1579 & 23 & -  \\ \hline \hline
 2D CNN \scriptsize{(our approach)} & 6 & 16 & 52 & 5 & 1 & 1\\
 \hline
\end{tabular}
}
}
\end{center}
\captionsetup{justification=centering, size=small}
\caption{Runtimes in seconds, rounded to the nearest integer. $^\star$without using node attributes}
\label{table:time}
\end{table*}

\vspace{-1cm}

\noindent \textbf{Discussion}. Replacing the raw counts by the empirical joint probability density function, either by normalizing the histograms, or with a Kernel Density Estimate, significantly deteriorated performance. This suggests that keeping the absolute counts is important, which makes sense, as some categories might be associated with larger or smaller graphs, on average. We also observed that increasing the number of channels to more than 5 does not yield better results. This was expected, as higher order channels contain less information than lower order channels. However, reducing the number of channels improves performance in some cases, probably because it plays a regularization role.

\begin{figure}[h]
  \centering
    \includegraphics[width=0.99\linewidth]{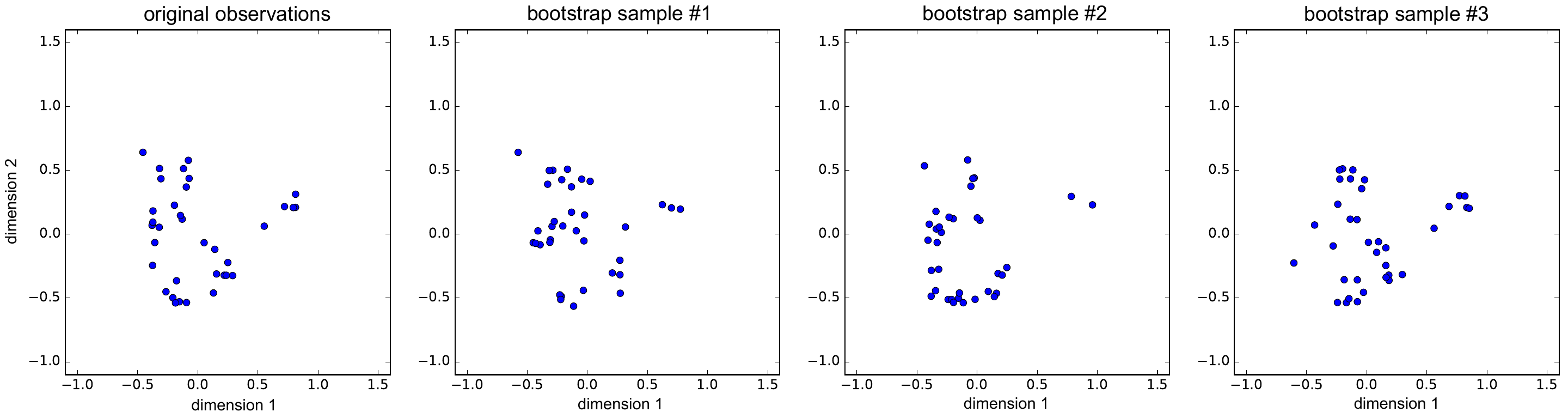}
\captionsetup{justification=justified,singlelinecheck=false, size=small}
    \caption{Example of data augmentation with the smoothed bootstrap generator, for graph ID \#3 of IMDB-B dataset (35 nodes, 133 edges).}
\label{fig:boot}
\end{figure}

\textit{Data augmentation}.
Generating more training examples by altering the input images is known to improve performance in image classification \cite{krizhevsky2012imagenet}. However, since we had direct access to the underlying data that were used to generate the images of the graphs, which is typically not the case in computer vision, we thought it would be more sensible to implement a synthetic data generation scheme at the \textit{node embeddings} level rather than at the image level. More precisely, we used a simple but powerful nonparametric technique known as the \textit{smoothed bootstrap with variance correction} \cite{silverman1986density}, detailed in Alg. \ref{alg:boot}. This generator is used in hydroclimatology to improve modeling of precipitation \cite{lall1996nonparametric} or streamflow \cite{sharma1997streamflow}. Unlike the traditional bootstrap \cite{efron1992bootstrap} which simply draws with replacement from the initial set of observations, the smoothed bootstrap can generate values outside of the original range while still being faithful to the structure of the underlying data \cite{rajagopalan1997multivariate,tixier2017construction}, as shown in Fig. \ref{fig:boot}.

\vspace{-0.25cm}

\vspace{-1.5cm}

\begin{algorithm}[h]
    \SetKwInOut{Input}{Input}
    \SetKwInOut{Output}{Output}
    \SetKw{Return}{return}
    \Input{list $L$ of $M$ arrays of original $d$-dimensional node embeddings, $p_{b} \in \mathbb{R}^{+}$}
    \Output{list $L'$ of $M' = \mathrm{int}\big(p_{b} \times M\big)$ arrays of synthetic node embeddings}
    $L'' \leftarrow $ select $M'$ elements from $L$ at random without replacement\\
    \For{$A \in \mathbb{R}^{|V_{A}| \times d}$ in $L''$}{
    $\mu_{A} \leftarrow$ $\mathrm{apply(mean,A)}$\\
    $\sigma^{2}_{A}\leftarrow$ $\mathrm{apply(var,A)}$ \\
    $h_{A} \leftarrow$ $\mathrm{apply(compute\_kde\_bandwidth,A)}$\\
        $A' \leftarrow \text{empty array}$\\
    \For{$j \leftarrow 1$ \KwTo $\mathrm{int}\Big(\mathcal{N}\big(|V_{A}|,\frac{|V_{A}|}{5}\big)$\Big)}{sample $i$ from $U(1,|V_{A}|)$ \\ $\epsilon \leftarrow$ $\mathrm{apply(\mathcal{N}(0,\sqrt{x}),h_{A})}$ \\
    \For{$k \leftarrow 1$ \KwTo $d$}{
    $A'[j,k] \leftarrow \mu_{A}[k] + \nicefrac{(A[i,k]-\mu_{A}[k] + \epsilon[k])}{\sqrt{1+\nicefrac{h_{A}[k]}{\sigma^{2}_{A}[k]}}}$
    }
    }
    \textbf{store} $A'$ in $L'$
    }
\Return{$L'$}
\caption{Smoothed bootstrap with variance correction \label{alg:boot}}
\end{algorithm}

\vspace{-0.8cm}

$|V_{A}|$ is the number of nodes in the graph whose node embeddings are contained in $A$. Note that $\mu_A,\sigma^{2}_{A}, h_A, \epsilon$ in lines 3, 4, 5 and 9 are all $d$-dimensional vectors (i.e., the functions are applied column-wise). $\mathcal{N}$ and $U$ in lines 7 and 8 are the normal and discrete uniform distributions. The function $\mathcal{N}(0,\sqrt{x})$ in line 9 is applied to each element of $h_A$. It generates Gaussian noise in each direction.
Line 5 computes the bandwidths of the Kernel Density Estimate (KDE) over each dimension. The bandwidth controls the degree of smoothing. While there are many different ways to evaluate the bandwidth, we used the classical and widespread Silverman's rule of thumb \cite{silverman1986density}, as shown in Eq. \ref{eq:silverman} below for a sample X of $N$ scalars:

\begin{equation} \label{eq:silverman}
\mathrm{silverman}\big(X\big) = \frac{0.9 \mathrm{min}\big(\sigma_{X},\frac{Q_{3}(X)-Q_{1}(X)}{1.34}\big)}{N^{\nicefrac{1}{5}}}
\end{equation}

Where $Q_{3}$ and $Q_{1}$ represent respectively the third and first quartiles, $\sigma_{X}$ is the standard deviation of the sample, and $N$ is the size of the sample.

Line 11 simulates a new embedding vector. Using the smoothed bootstrap scheme can be viewed as sampling from the KDE. This is consistent with our way of representing graphs as images, since a KDE is nothing more than a smoothed histogram.

In our experiments, using the smoothed bootstrap improved performance only on REDDIT-B ($+0.33\%$ in accuracy), for $p_b=0.2$ (i.e., augmenting the dataset with 20\% of synthetic graphs). Other values of $p_b$ (0.05, 0.1, 0.5) were not successful. Further research is thus necessary to understand how to make the proposed data augmentation strategy more effective.

\textit{Limitations}. Even though results are very good out-of-the-box in most cases, finding an embedding algorithm that works well, or the right combination of parameters for a given dataset, can require some efforts. For instance, on COLLAB and IMDB-B (the only two datasets on which we do not reach best performance), we hypothesize that our results are inferior because the default parameter values of \texttt{node2vec} may not be well-suited to very dense graphs such as the ones found in COLLAB and IMDB-B (diameter$<2$, density $>50$). Optimizing the \texttt{node2vec} parameters on these datasets probably requires more than a coarse grid search.

\section{Related work}
Motivated by the outstanding performance recently reached by CNNs in computer vision, e.g. \cite{vinyals2015show,krizhevsky2012imagenet}, much research has been devoted to generalizing CNNs to graphs.
Solutions fall into two broad categories: \textit{spatial} and \textit{spectral} techniques \cite{bruna2013spectral}. Spectral approaches \cite{defferrard2016convolutional,kipf2016semi} invoke the convolution theorem from signal processing theory to perform graph convolutions as pointwise multiplications in the Fourier domain of the graph. The basis used to send the graph to the Fourier domain is given by the SVD decomposition of the Laplacian matrix of the graph, whose eigenvalues can be viewed as frequencies. By contrast, spatial methods \cite{zhang2018end,niepert2016learning,vialatte2016generalizing} operate directly on the graph structure. For instance, in \cite{niepert2016learning}, the algorithm first determines the sequence of nodes for which neighborhood graphs (of equal size) are created. To serve as receptive fields, the neighborhood graphs are then normalized, i.e., mapped to a vector space with a linear order, in which nodes with similar structural roles in the neighborhood graphs are close to each other. Normalization is the central step, and is performed via a labeling procedure. A 1D CNN architecture is finally applied to the receptive fields.
While the aforementioned sophisticated frameworks have made great strides, we showed in this paper that graphs can also be processed by vanilla 2D CNN architectures.

\vspace{-0.25cm}

\section{Conclusion}
We showed that CNN for images can be used for learning graphs in a completely off-the-shelf manner. Our approach is flexible and can take continuous node attributes into account. We reach better results than state-of-the-art graph kernels and graph CNN models on 4 real-world datasets out of 6. Furthermore, these good results were obtained with limited parameter tuning and by using a basic 2D CNN model. From a time complexity perspective, our approach is preferable to graph kernels too, allowing to process bigger datasets featuring larger graphs.

\section{Acknowledgments}
We thank the anonymous reviewers for their helpful comments. The GPU used in this project was donated by NVidia as part of their GPU grant program.

%
%
%
\bibliographystyle{splncs04}
\bibliography{main}

\begin{thebibliography}{10}
\providecommand{\url}[1]{\texttt{#1}}
\providecommand{\urlprefix}{URL }
\providecommand{\doi}[1]{https://doi.org/#1}

\bibitem{abadi2016tensorflow}
Abadi, M., Agarwal, A., Barham, P., Brevdo, E., Chen, Z., Citro, C., Corrado,
  G.S., Davis, A., Dean, J., Devin, M., et~al.: Tensorflow: Large-scale machine
  learning on heterogeneous distributed systems. arXiv preprint
  arXiv:1603.04467  (2016)

\bibitem{borgwardt2005shortest}
Borgwardt, K.M., Kriegel, H.: Shortest-path kernels on graphs. In: Proceedings
  of the 5th International Conference on Data Mining. pp. 74--81 (2005)

\bibitem{borgwardt2005protein}
Borgwardt, K.M., Ong, C.S., Sch{\"o}nauer, S., Vishwanathan, S., Smola, A.J.,
  Kriegel, H.P.: Protein function prediction via graph kernels. Bioinformatics
  \textbf{21}(suppl 1),  i47--i56 (2005)

\bibitem{bottou2007support}
Bottou, L., Lin, C.J.: Support vector machine solvers. Large scale kernel
  machines  \textbf{3}(1),  301--320 (2007)

\bibitem{bruna2013spectral}
Bruna, J., Zaremba, W., Szlam, A., LeCun, Y.: Spectral networks and locally
  connected networks on graphs. arXiv preprint arXiv:1312.6203  (2013)

\bibitem{chollet2015keras}
Chollet, F., et~al.: Keras. \url{https://github.com/fchollet/keras} (2015)

\bibitem{cortes1995support}
Cortes, C., Vapnik, V.: Support-vector networks. Machine learning
  \textbf{20}(3),  273--297 (1995)

\bibitem{defferrard2016convolutional}
Defferrard, M., Bresson, X., Vandergheynst, P.: Convolutional neural networks
  on graphs with fast localized spectral filtering. In: Advances in Neural
  Information Processing Systems. pp. 3837--3845 (2016)

\bibitem{efron1992bootstrap}
Efron, B.: Bootstrap methods: another look at the jackknife. In: Breakthroughs
  in Statistics, pp. 569--593. Springer (1992)

\bibitem{feragen2013scalable}
Feragen, A., Kasenburg, N., Petersen, J., de~Bruijne, M., Borgwardt, K.:
  Scalable kernels for graphs with continuous attributes. In: Advances in
  Neural Information Processing Systems. pp. 216--224 (2013)

\bibitem{gartner2003graph}
G{\"a}rtner, T., Flach, P., Wrobel, S.: On graph kernels: Hardness results and
  efficient alternatives. In: Learning Theory and Kernel Machines, pp.
  129--143. Springer (2003)

\bibitem{glorot2010understanding}
Glorot, X., Bengio, Y.: Understanding the difficulty of training deep
  feedforward neural networks. In: Aistats. vol.~9, pp. 249--256 (2010)

\bibitem{grover2016node2vec}
Grover, A., Leskovec, J.: node2vec: Scalable feature learning for networks. In:
  Proceedings of the 22nd ACM SIGKDD international conference on Knowledge
  discovery and data mining. pp. 855--864. ACM (2016)

\bibitem{KKMMN2016}
Kersting, K., Kriege, N.M., Morris, C., Mutzel, P., Neumann, M.: Benchmark data
  sets for graph kernels (2016), \url{http://graphkernels.cs.tu-dortmund.de}

\bibitem{kingma2014adam}
Kingma, D., Ba, J.: Adam: A method for stochastic optimization. arXiv preprint
  arXiv:1412.6980  (2014)

\bibitem{kipf2016semi}
Kipf, T.N., Welling, M.: Semi-supervised classification with graph
  convolutional networks. arXiv preprint arXiv:1609.02907  (2016)

\bibitem{krizhevsky2012imagenet}
Krizhevsky, A., Sutskever, I., Hinton, G.E.: Imagenet classification with deep
  convolutional neural networks. In: Advances in neural information processing
  systems. pp. 1097--1105 (2012)

\bibitem{lall1996nonparametric}
Lall, U., Rajagopalan, B., Tarboton, D.G.: A nonparametric wet/dry spell model
  for resampling daily precipitation. Water resources research  \textbf{32}(9),
   2803--2823 (1996)

\bibitem{lecun1998gradient}
LeCun, Y., Bottou, L., Bengio, Y., Haffner, P.: Gradient-based learning applied
  to document recognition. Proceedings of the IEEE  \textbf{86}(11),
  2278--2324 (1998)

\bibitem{mikolov2013efficient}
Mikolov, T., Chen, K., Corrado, G., Dean, J.: Efficient estimation of word
  representations in vector space. arXiv preprint arXiv:1301.3781  (2013)

\bibitem{morris2016faster}
Morris, C., Kriege, N.M., Kersting, K., Mutzel, P.: Faster kernels for graphs
  with continuous attributes via hashing. In: Data Mining (ICDM), 2016 IEEE
  16th International Conference on. pp. 1095--1100. IEEE (2016)

\bibitem{neumann2012efficient}
Neumann, M., Patricia, N., Garnett, R., Kersting, K.: Efficient graph kernels
  by randomization. In: Joint European Conference on Machine Learning and
  Knowledge Discovery in Databases. pp. 378--393. Springer (2012)

\bibitem{niepert2016learning}
Niepert, M., Ahmed, M., Kutzkov, K.: Learning convolutional neural networks for
  graphs. In: Proceedings of the 33rd annual international conference on
  machine learning. ACM (2016)

\bibitem{nikolentzos2017matching}
Nikolentzos, G., Meladianos, P., Vazirgiannis, M.: Matching node embeddings for
  graph similarity. In: AAAI. pp. 2429--2435 (2017)

\bibitem{orsini2015graph}
Orsini, F., Frasconi, P., De~Raedt, L.: Graph invariant kernels. In:
  Proceedings of the Twenty-fourth International Joint Conference on Artificial
  Intelligence. pp. 3756--3762 (2015)

\bibitem{scikitlearn}
Pedregosa, F., Varoquaux, G., Gramfort, A., Michel, V., Thirion, B., Grisel,
  O., Blondel, M., Prettenhofer, P., Weiss, R., Dubourg, V., Vanderplas, J.,
  Passos, A., Cournapeau, D., Brucher, M., Perrot, M., Duchesnay, E.:
  Scikit-learn: Machine learning in python. Journal of Machine Learning
  Research  \textbf{12},  2825--2830 (2011)

\bibitem{perozzi2014deepwalk}
Perozzi, B., Al-Rfou, R., Skiena, S.: Deepwalk: Online learning of social
  representations. In: Proceedings of the 20th ACM SIGKDD international
  conference on Knowledge discovery and data mining. pp. 701--710. ACM (2014)

\bibitem{rajagopalan1997multivariate}
Rajagopalan, B., Lall, U., Tarboton, D.G., Bowles, D.: Multivariate
  nonparametric resampling scheme for generation of daily weather variables.
  Stochastic Hydrology and Hydraulics  \textbf{11}(1),  65--93 (1997)

\bibitem{sharma1997streamflow}
Sharma, A., Tarboton, D.G., Lall, U.: Streamflow simulation: A nonparametric
  approach. Water resources research  \textbf{33}(2),  291--308 (1997)

\bibitem{shervashidze2009efficient}
Shervashidze, N., Petri, T., Mehlhorn, K., Borgwardt, K.M., Vishwanathan, S.:
  {Efficient Graphlet Kernels for Large Graph Comparison}. In: Proceedings of
  the International Conference on Artificial Intelligence and Statistics. pp.
  488--495 (2009)

\bibitem{shervashidze2011weisfeiler}
Shervashidze, N., Schweitzer, P., Van~Leeuwen, E.J., Mehlhorn, K., Borgwardt,
  K.M.: {Weisfeiler-Lehman Graph Kernels}. The Journal of Machine Learning
  Research  \textbf{12},  2539--2561 (2011)

\bibitem{silverman1986density}
Silverman, B.W.: Density estimation for statistics and data analysis, vol.~26.
  CRC press (1986)

\bibitem{srivastava2014dropout}
Srivastava, N., Hinton, G.E., Krizhevsky, A., Sutskever, I., Salakhutdinov, R.:
  Dropout: a simple way to prevent neural networks from overfitting. Journal of
  Machine Learning Research  \textbf{15}(1),  1929--1958 (2014)

\bibitem{tixier2017construction}
Tixier, A.J.P., Hallowell, M.R., Rajagopalan, B.: Construction safety risk
  modeling and simulation. Risk analysis  (2017)

\bibitem{tobler1970computer}
Tobler, W.R.: A computer movie simulating urban growth in the detroit region.
  Economic geography  \textbf{46}(sup1),  234--240 (1970)

\bibitem{vialatte2016generalizing}
Vialatte, J.C., Gripon, V., Mercier, G.: Generalizing the convolution operator
  to extend cnns to irregular domains. arXiv preprint arXiv:1606.01166  (2016)

\bibitem{vinyals2015show}
Vinyals, O., Toshev, A., Bengio, S., Erhan, D.: Show and tell: A neural image
  caption generator. In: Proceedings of the IEEE Conference on Computer Vision
  and Pattern Recognition. pp. 3156--3164 (2015)

\bibitem{yanardag2015deep}
Yanardag, P., Vishwanathan, S.: Deep graph kernels. In: Proceedings of the 21th
  ACM SIGKDD International Conference on Knowledge Discovery and Data Mining.
  pp. 1365--1374. ACM (2015)

\bibitem{zhang2018end}
Zhang, M., Cui, Z., Neumann, M., Chen, Y.: An end-to-end deep learning
  architecture for graph classification  (2018)

\end{thebibliography}

\end{document}